\documentclass[letterpaper, 10 pt, conference]{ieeeconf}
\IEEEoverridecommandlockouts
\usepackage{amsmath,amsfonts}
\usepackage[ruled]{algorithm2e}
\usepackage{array}
\usepackage{textcomp}
\usepackage{stfloats}
\usepackage{url}
\usepackage{verbatim}
\usepackage{graphicx}
\usepackage{cite}
\usepackage[utf8]{inputenc}
\usepackage[english]{babel}
\usepackage[dvipsnames]{xcolor}
\usepackage{xspace}
\usepackage[colorlinks=true,allcolors=blue,urlcolor=black]{hyperref} 

\newcommand{\etal}{et~al.~}

\newcommand{\eg}{e.g.~}

\usepackage{multirow}
\usepackage{url}
\usepackage{comment}
\usepackage{caption}
\usepackage{subcaption}
\usepackage{booktabs}

\begin{document}

\bstctlcite{IEEEexample:BSTcontrol} 

\title{Doppler-only Single-scan 3D Vehicle Odometry}

\author{Andres Galeote-Luque, Vladimír Kubelka, Martin Magnusson,\\Jose-Raul Ruiz-Sarmiento and Javier Gonzalez-Jimenez
\thanks{*This work has been supported by the grant program PRE2018-085026 and the research project ARPEGGIO (PID2020-117057GB-I00), all funded by the Spanish Government. The authors would also like to express their gratitude to Maximilian Hilger for his assistance with the experimental part of this work.}
\thanks{Andres Galeote-Luque (corresponding author), Jose-Raul Ruiz-Sarmiento and Javier Gonzalez-Jimenez ({\tt\small{andresgalu}; \tt\small{jotaraul}; \tt\small{javiergonzalez} \tt\small{@uma.es}}) are with Machine Perception and Intelligent Robotics (MAPIR) Group, System Engineering and Automation Department, University of Malaga, Spain.}%
\thanks{Vladimír Kubelka and Martin Magnusson ({\tt\small{vladimir.kubelka}; \tt\small{martin.magnusson} \tt\small{@oru.se}}) are with the Mobile Robotics \& Olfaction (MRO), AASS research centre, Örebro University, Sweden. }%
}

\maketitle

\begin{abstract}
We present a novel 3D odometry method that recovers the full motion of a vehicle only from a Doppler-capable range sensor.
It leverages the radial velocities measured from the scene, estimating the sensor's velocity from a single scan.  The vehicle's 3D motion, defined by its linear and angular velocities, is calculated taking into consideration its kinematic model which provides a constraint between the velocity measured at the sensor frame and the vehicle frame. 

Experiments carried out prove the viability of our single-sensor method compared to mounting an additional IMU. 
Our method provides the translation of the sensor, which cannot be reliably determined from an IMU, as well as its rotation.
Its short-term accuracy and fast operation ($\sim$5ms) make it a proper candidate to supply the initialization to more complex localization algorithms or mapping pipelines. Not only does it reduce the error of the mapper, but it does so at a comparable level of accuracy as an IMU would. All without the need to mount and calibrate an extra sensor on the vehicle.
\end{abstract}


\section{INTRODUCTION}
Self-localization is one of the fundamental components of autonomous mobile robots and vehicles.
This problem can be tackled by employing different sensors, but cameras and lidars are among the most widely used in contemporary methods.
These sensors provide enough advantages to justify their relevance, though their performance can be severely hindered under challenging circumstances, like extreme lighting or weather conditions, or the presence of dust or mist \cite{carballo2020libre}. 

Radar sensors, on the other hand, are a promising but underexplored alternative able to provide geometric information of its surroundings even in challenging low-visibility scenarios \cite{mielle2019comparative, venon2022millimeter, hong2022radarslam}. This makes them highly suitable for underground operations, construction sites, and other harsh environments. 
The relevance of radars in industrial and on-road applications has driven sensor technology forward, making them more affordable, accurate, and fast.
Among the various advances, it is worth highlighting the introduction of radar sensors measuring in 3D  (azimuth, elevation and range), as well as their ability to estimate the radial velocity of each sensed 3D point.
The latter is possible by leveraging precise phase measurements of the returning signal,
which allows the sensor to measure the velocity of a point along the radar-point direction. Further on, we denote it as Doppler velocity.
Note that this technology has also been implemented in certain lidars, so from now on we will refer to all of them as Doppler-capable range sensors.

\begin{figure}
    \centering
    \includegraphics[angle=90, width=0.92\linewidth]{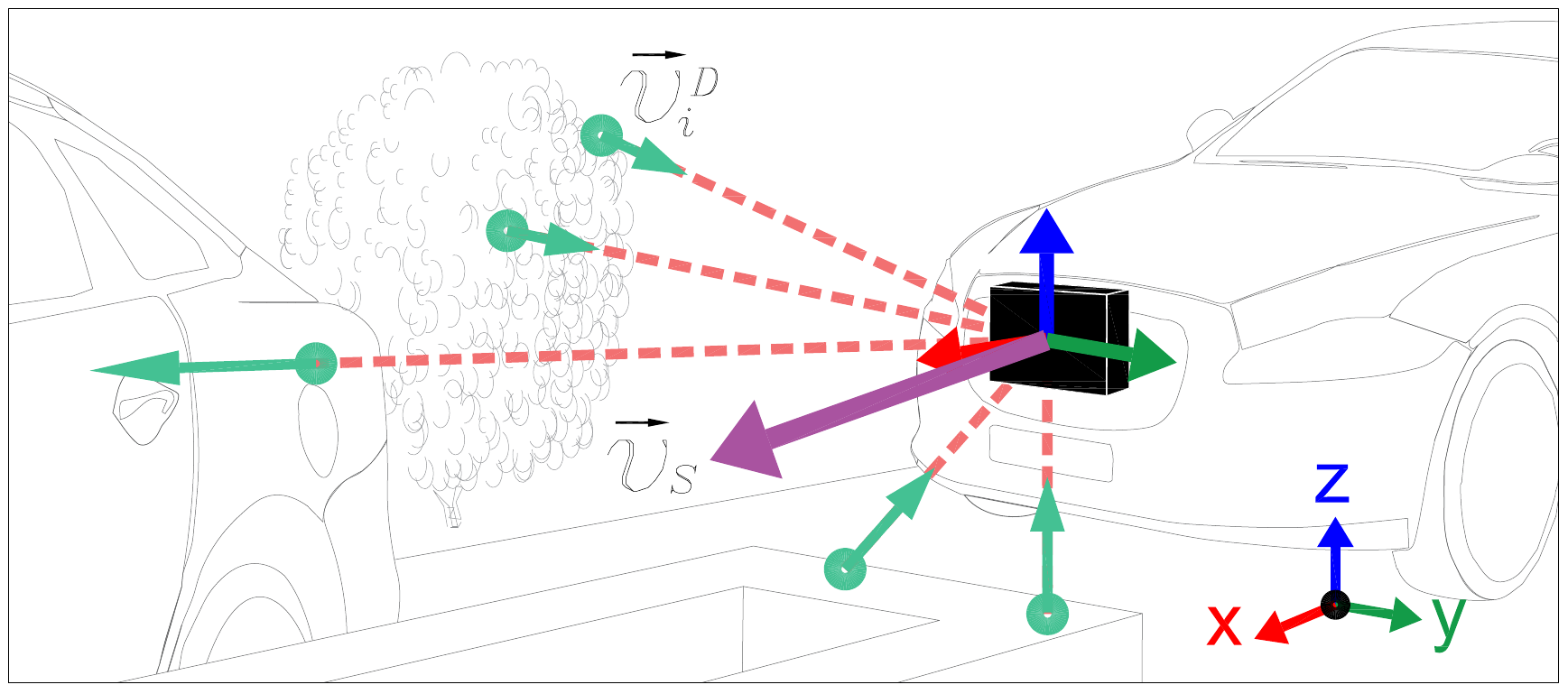}

    \includegraphics[width=0.65\linewidth]{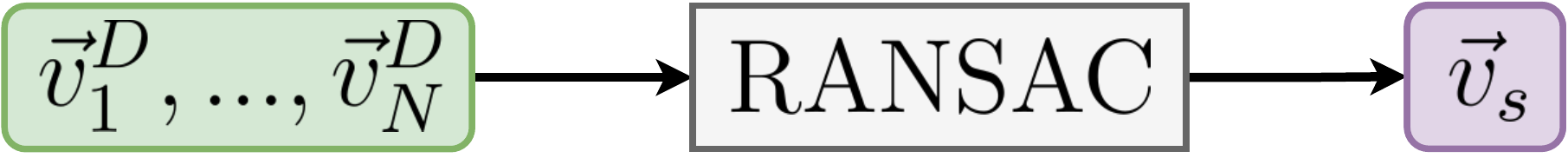}

    \vspace{5px}
    
    \includegraphics[angle=90, width=0.92\linewidth]{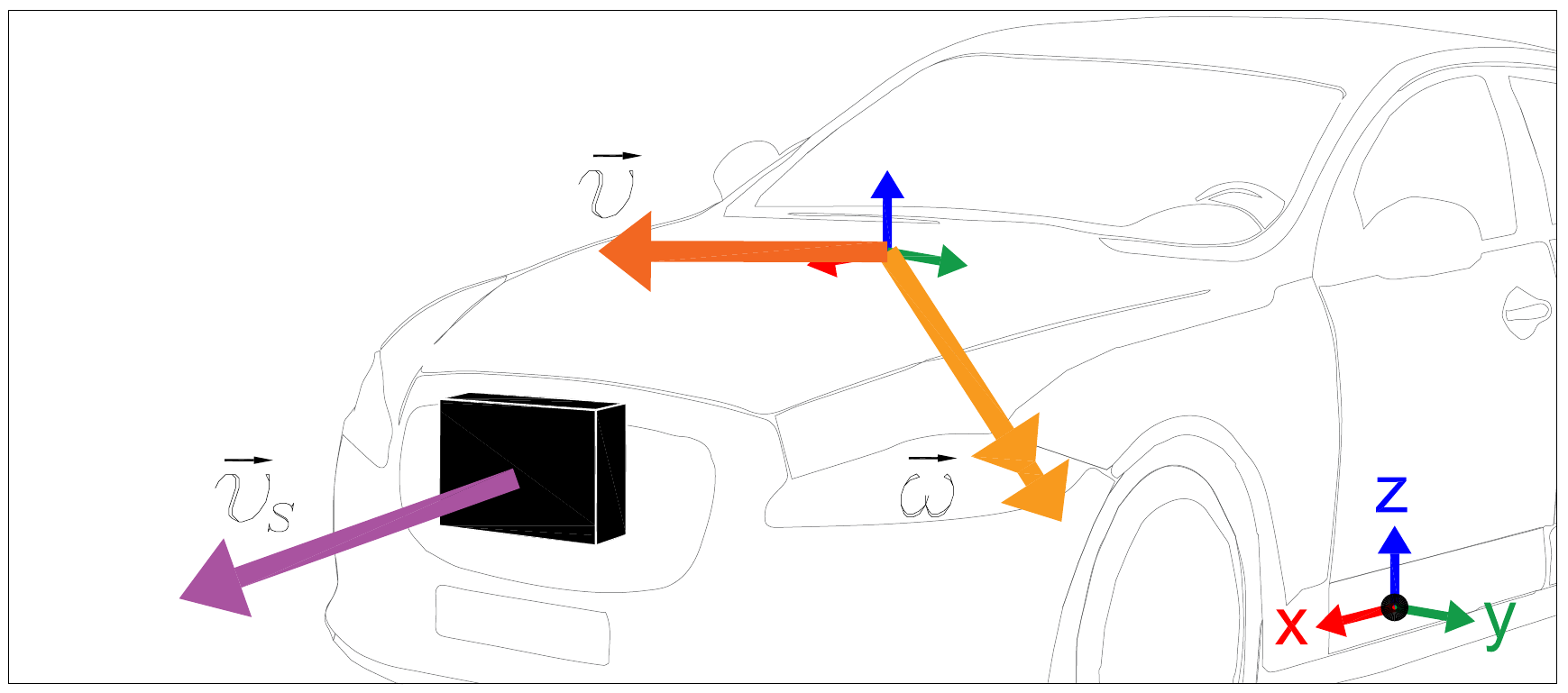}

    \includegraphics[width=0.65\linewidth]{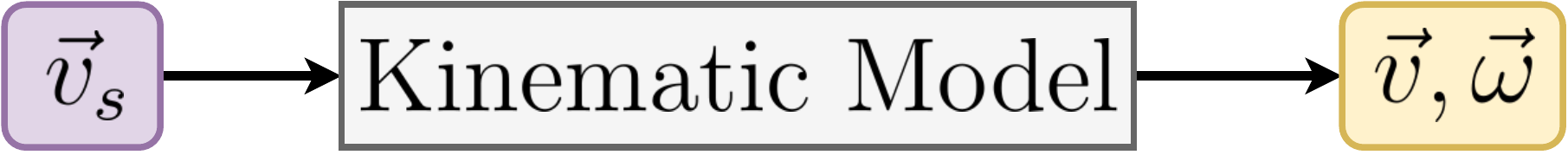}

    \caption{
    Representation of the working principle of the proposed method. Top, the radial velocity $\vec{v}_i^D$ of the observed points in the scene is leveraged to estimate the sensor's linear velocity $\vec{v}_s$. 
    Bottom, the kinematic model provides the relation between $\vec{v}_s$ and both the linear $\vec{v}$ and angular $\vec{\omega}$ velocities of the vehicle.
    }
    \label{fig:intro}
\end{figure}

The radial velocity
provided by these sensors has proven to be advantageous for odometry methods, aiding in the segmentation of dynamic and static objects \cite{kingery2022improving}, as well as introducing more constraints to the movement estimation \cite{hexsel2022dicp}.
Given the right conditions, it can be as powerful as to recover the 2D ego-motion of the sensor from a single scan of its surroundings \cite{kellner2013instantaneous, kellner2014instantaneous}.
Note that, to estimate the localization, these works rely upon either knowing the kinematic model of the vehicle or mounting multiple sensors, since not all three degrees of freedom (DOFs) of a 2D movement are observable with just the radial velocity of points of the scene.
Estimating the movement of the sensor this way removes the need to perform data association, bringing it closer to employing an Inertial Measurement Unit (IMU) or wheel odometry.
These methods provide the relative motion of the sensor, and tend to accumulate error over time, also known as \textit{drift}.
However, their short-term accuracy obtained in a fast and simple way makes them widely adopted in different odometry solutions.

In this paper, we propose a 3D odometry method that relies only on the Doppler velocity to estimate the vehicle motion from a single scan, hence avoiding data association.
Doppler-capable range sensors can only capture the radial velocity of a point.
Assuming most of the scene is static, the linear velocity of the sensor can be recovered via RANSAC \cite{ransac}. 
However, there is a lack of information to obtain the 6 DOFs of a 3D movement.
The observability of this problem was studied by Yoon \etal \cite{yoon2023need}, where a gyroscope was added to completely determine the vehicle 3D motion. 
However, in this work we propose a simpler solution closer to the one presented by Kellner \cite{kellner2013instantaneous} but in 3D: by taking into consideration the kinematic model of the vehicle, we can recover its 3D movement using only the Doppler velocities.
In the following sections, we describe the proposed odometry algorithm, as well as the experiments performed to validate it. An overview of its working principles can be seen in Figure \ref{fig:intro}.
The results obtained prove that the presented method yields a short-term accurate motion estimation.
When it is employed as initialization of more advanced localization algorithms like mapper pipelines, it improves the accuracy without needing a different sensor mounted on the vehicle.
The code is publicly available at \href{https://github.com/andresgalu/doppler_odometry}{https://github.com/andresgalu/doppler\_odometry}.

\section{RELATED WORK}

The first publication to recover the ego-motion of a vehicle employing Doppler velocity without data association is due to Yokoo \etal{}~\cite{yokoo2009indoor}, who employ 1D radar sensors mounted on a vehicle with Ackermann steering performing 2D planar motion. They propose two sensor configurations to recover the forward and angular velocities, one with two radars mounted in front of the vehicle, and another with a single radar and a gyroscopic sensor.
The method requires the observed objects to be static in order to provide a reliable localization estimation.
Kellner \etal expanded this concept for a single 2D radar \cite{kellner2013instantaneous}. Since the sensor can only measure the radial velocity of the objects, 
the angular velocity cannot be directly observed.
To circumvent this issue, again the vehicle has to comply with the Ackermann kinematic model, reducing the DOFs from 3 (complete 2D motion) to 2 (forward and angular velocities).
The increase in data points due to leveraging a 2D sensor removes the need to assume a static scene. By applying RANSAC \cite{ransac}, dynamic points can be identified and discarded as outliers.
Kellner \etal later developed a method to completely recover the 3 DOFs of the 2D motion by employing multiple radar sensors \cite{kellner2014instantaneous}, avoiding the Ackermann model requirement.

Stepping into 3D movement, works of both Kramer \etal \cite{kramer2020radar} and Doer and Trommer \cite{DoerMFI2020} combine the information from a 3D radar and an IMU to recover the 6 DOFs of the motion, employing batch optimization by the former and an Extended Kalman Filter (EKF) by the latter.
The use of an IMU as a way to make the movement observable is reminiscent of the second configuration proposed by Yokoo \etal \cite{yokoo2009indoor}.
Similarly, Yoon \etal \cite{yoon2023need} employ a Doppler-capable 3D lidar and a gyroscope to decouple the 3D motion estimation,
with the lidar and gyroscope yielding the translation and rotation respectively. Said work includes a study of the observability of the motion, explaining the need for either a gyroscope or multiple sensors to completely recover the 3D movement.
Even without a 3D radar, Park \etal \cite{park20213d} estimate the 3D motion of a vehicle by compositing two orthogonal 2D radar sensors, along with an IMU.

As well as providing a motion estimation, the Doppler velocity has proven to be helpful in more traditional odometry algorithms that perform data association, typically registering two consecutive point clouds.
The most common approach to use the Doppler velocity is to add a cost function to the optimization problem, where the measured radial velocity is compared to the expected velocity of a point given an estimation of the motion. 
This can be applied to registration problems commonly found in the literature such as Iterative Closest Point (ICP) \cite{icp} variations \cite{retan2021radar, hexsel2022dicp, wu2022picking}, Normal Distribution Transform \cite{ndt} (NDT) \cite{rapp2017probabilistic}, and similar approaches \cite{barjenbruch2015joint}.
The work of Monaco and Brennan \cite{monaco2020radarodo} stands out for decoupling the motion estimation: the translation is obtained from the Doppler velocity, while the collected spatial information provides the rotation.
The recent rise of Neural Networks (NN) has also reached radar odometry as can be seen in the work of Rennie \etal~\cite{rennie2023doppler}, where Doppler velocity is used to obtain an estimation of the motion which can later be fused with the scan registration result.

In contrast to the literature that is reviewed above, the method proposed in this article provides the 3D motion of a Doppler-capable range sensor by only leveraging the radial velocities from the scene. By taking into consideration the kinematic model of the vehicle the movement can be recovered without needing an extra IMU.
The efficiency and short-term accuracy of this algorithm make it more than helpful to provide an estimation of the motion, which is also valuable as initialization in more resource-hungry algorithms.

\section{METHOD OVERVIEW}
\label{sec:overview}
In this section, we explain how the proposed method estimates the 3D movement of the vehicle from the Doppler velocity measurements of points in the scene.
The method first estimates the sensor velocity and then finds the vehicle movement that fulfills a certain kinematic model and explains the observed sensor velocity.
In Section \ref{sec:sensorvel} we will describe how the velocity of the sensor itself can be calculated from the observed radial velocities. 
As explained before, from the sensor velocity we cannot obtain the 6 DOFs of the vehicle's 3D motion. By introducing the kinematic model of the vehicle in Section \ref{sec:kinematic}, the DOFs of the movement get reduced and hence it can be solved. 
Finally, Section \ref{sec:covariance} will analyze the variance of the estimated vehicle movement variables.

\subsection{SENSOR VELOCITY}
\label{sec:sensorvel}

In this section, we will delve into the procedure that yields the sensor velocity from the Doppler velocities of the observed points. First, we will derive the equations assuming the scene is completely static, and then we will tackle the issue of removing dynamic objects.

Assuming a static scene, the velocity of an observed object is the opposite of the sensor velocity $v_s$.
Since the Doppler velocity from each point $v^D_i$ is only measured along the radial direction, this value is the result of projecting the opposite of the sensor velocity on the radial direction.
\begin{equation}
\label{eq:doppler}
    -v^D_i = 
    \left[ \begin{matrix}
    \cos\phi_i \cos\theta_i & \sin\phi_i \cos\theta_i & \sin\theta_i
    \end{matrix} \right]
    \left[ \begin{matrix}
    v_{sx} \\ v_{sy} \\ v_{sz}
    \end{matrix} \right]
\end{equation}

The azimuth $\phi$ and elevation angles $\theta$ of each point $i$ are obtained by converting their cartesian coordinates to spherical.
Applying Equation \eqref{eq:doppler} to all sensed points results in a set of linear equations in the form of $B=Ax$ that can be solved with least squares regression.
To reduce the influence of noise in the final result, we weight each point by its signal power. The sensor velocity is thus estimated as follows, with $W$ being the diagonal matrix of the weights.
\begin{equation}
\label{eq:lsq}
    v_s = (A^\top WA)^{-1}(A^\top WB)
\end{equation}

In the case there are dynamic objects in the scene, not all the points will follow the same model. 
In our implementation, we chose RANSAC \cite{ransac} to get rid of these outlier points that do not comply with the estimated sensor velocity. 
They are labeled as dynamic, which can be helpful for later processing of the data.

\subsection{VEHICLE KINEMATIC MODEL}
\label{sec:kinematic}

\begin{figure*}
    \centering
    \includegraphics[width=0.95\linewidth]{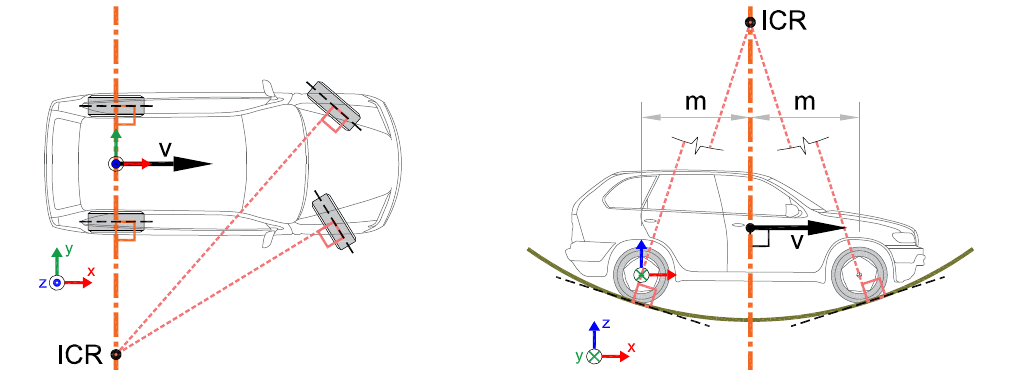}
    \begin{subfigure}{0.49\textwidth}
        \centering
        \caption{Top view of the vehicle movement on the XY plane.}
        \label{fig:ackerman}
    \end{subfigure}
    \hfill
    \begin{subfigure}{0.49\textwidth}
        \centering
        \caption{Side view of the vehicle movement on the XZ plane.}
        \label{fig:hill}
    \end{subfigure}
    \caption{Simple cases of movement of the vehicle showing the location of the line along which the ICR can be found (orange). The velocity of points located on the ICR axis is perpendicular to it.}
    \label{fig:model}
\end{figure*}

Now that the sensor velocity has been calculated from the Doppler velocity measurements, we can estimate the vehicle movement. 
There exists an infinite number of 3D motions that explain the calculated sensor velocity, and we lack the information to determine all the 6 DOFs.
However, by constraining the vehicle motion to a certain kinematic model, it is possible to simplify the movement enough to obtain an estimation.

First of all, we need to define the frame of reference used throughout this section. Its position in the vehicle is not particularly relevant, but its orientation is.
As shown in Figure \ref{fig:model} we establish the Y axis to be aligned with the rear wheel axis, pointing to the left of the vehicle. The X axis is perpendicular to it and points toward the front wheel axis. Finally, the Z axis points upwards, perpendicular to both X and Y.
Without loss of generality, we will locate the frame of reference at the center of the rear wheel axis.

For the sake of clarity, we will start analyzing the movement of an Ackermann steering vehicle moving along a flat surface, which is simply a 2D motion.
In that case, the Instant Center of Rotation (ICR) can be found somewhere along the line 
of contact between the rear wheels and the ground, but it can be well approximated as the line that goes through the rear wheel axis.
From now on we will refer to this line as the Z-ICR axis, since it relates to the rotation around Z.
The velocity along the Y direction of any point located in the Z-ICR axis is null by definition.
See Figure \ref{fig:ackerman} for a graphic representation.
Note that the key concept here is to have the ICR restricted to a given line, and the Ackermann kinematic model is just an example. As long as the kinematic model fulfills this requirement, like in differential drive and skid-steer vehicles, the viability of the proposed method holds.

For a 3D motion, the movement along Z has to be analyzed too. Similar to the previous paragraph, we can first study the simplified movement where the vehicle only moves in the XZ plane, along a curved surface.
The ICR in this case is located at the intersection of the lines perpendicular to the contact point between the wheels and the ground.
Assuming the curvature of the ground is locally constant in the relatively short distance between both wheel axes, the ICR will be located in a line parallel to the Z axis that goes through the midpoint between both wheel axes.
Every point located along this Y-ICR axis will consequently have zero velocity in the Z direction.
Figure \ref{fig:hill} shows this concept for better clarity.
This holds true even when driving on a flat plane, in which case every point in the vehicle has zero vertical velocity.

Until now, we have studied the rotation around the Z axis (yaw) and Y axis (pitch) through two basic movement examples (Figures \ref{fig:ackerman} and \ref{fig:hill} respectively).
The roll, or rotation around the X axis, can be assumed negligible for a vehicle moving on a road.
The complete 3D motion can then be seen as a combination of these two basic movements, and the introduced restrictions will still be present in the resulting 3D motion model. Namely, the velocity of points located in the Z-ICR axis will have no Y component, and similarly points in the Y-ICR (the middle of the vehicle along the X axis) will have zero Z velocity.

We can now recover the motion of the vehicle by applying the mentioned restrictions, knowing that the velocities of two points $\vec{p}, \vec{s}$ of the same rigid moving object are related through their relative position and angular velocity $\vec{\omega}$:
\begin{equation}
\label{eq:rigid_vel}
    \vec{v}_p = \vec{v}_s + \vec{\omega} \times (\vec{p} - \vec{s})
\end{equation}

Using the previously estimated sensor velocity $\vec{v}_s$ \eqref{eq:lsq}, along with the restrictions, we can recover the angular velocity vector. To that end, point $\vec{s}$ is now the sensor position, and point $\vec{p}$ belongs to the Z-ICR axis, fulfilling the restriction of having zero Y velocity. Since all points on said axis comply with the requirement, we choose a point $\vec{p}_a$ located at the same Y coordinate as the sensor, for simplicity.
\begin{equation}
    \begin{cases}   
    \vec{p}_a = \left[ 0, s_y, 0 \right]^\top \\ 
    \vec{v}_{a} = \left[ v_{ax}, 0, v_{az} \right]^\top
    \end{cases} 
    \rightarrow \quad
    0 = v_{sy} - \omega_z s_x + \omega_x s_z
\end{equation}
%
In a similar manner, we can apply the second restriction, which indicates that a point $\vec{p}_b$ located on the Y-ICR (in the middle of the vehicle) has zero Z velocity.
\begin{equation}
    \begin{cases} 
    \vec{p}_b = \left[ m, 0, s_z \right]^\top \\ 
    \vec{v}_b = \left[ v_{bx}, v_{by}, 0 \right]^\top
    \end{cases}
    \rightarrow \quad
    0 = v_{sz} - \omega_x s_y - \omega_y (m - s_x)
\end{equation}
Where $m$ is half the distance between the two wheel axes of the vehicle, see Figure \ref{fig:hill}.
Finally, we apply the restriction ${\omega_x=0}$ since the rotation around the X axis is negligible, and thus the angular velocity vector can be recovered.
\begin{equation}
\label{eq:angvel}
    \vec{\omega} = 
    \left[ \begin{matrix}
        0 &
        v_{sz}/(m - s_x) &
        v_{sy}/s_x
    \end{matrix} \right]^\top
\end{equation}

The global pose of the sensor with respect to the reference frame of the world can be updated from the previous instance after a period of time $\Delta t$ by employing the velocity $\vec{v}_s$ for the position $\vec{p}_s$, and the angular velocity $\vec{\omega}$ for the orientation $R$.
Note that the velocity of any point of the vehicle can be calculated by applying (\ref{eq:rigid_vel}), and thus its pose can be updated:
\begin{equation}
\begin{cases} 
    R(t+\Delta t) &=\quad R(t) e^{\vec{\omega} \Delta t} \\ 
    \vec{p}_s(t+\Delta t) &=\quad \vec{p}_s(t) + R(t)\vec{v}_s \Delta t
    \end{cases}    
\end{equation}

\subsection{COVARIANCE ANALYSIS}
\label{sec:covariance}

In this section, we analyze the variance of the estimated sensor and angular velocities, in order to better understand how their accuracy can be improved.
First, we start with the sensor velocity $\vec{v}_s$ estimated in Section \ref{sec:sensorvel}. Since we employed least squares (\ref{eq:lsq}) to find the velocity, its covariance matrix $C_v$ can be estimated from the residuals $\rho = A \vec{v}_s - B$ \cite{hastie2009elements}, with $N$ being the rows of $A$:
\begin{equation}
\label{eq:cvfinal}
    C_v = \frac{\rho^\top W\rho}{N-3} (A^\top WA)^{-1}
\end{equation}

Other than the noise of the measurements, the distribution of the points in the scene is highly significant. The more spread they are, the higher the accuracy of the sensor velocity will be.
The covariance matrix $C_\omega$ of the angular velocity vector can then be calculated via error propagation:
\begin{equation}
\label{eq:covomega}
    C_\omega = J C_v J^\top = \left[ \begin{matrix}
        0 & 0 & 0 \\
        0 & \sigma_{vz}^2/(m-s_x)^2 & \sigma_{vyz}^2/(s_xm - s_x^2) \\
        0 & \sigma_{vyz}^2/(s_xm-s_x^2) & \sigma_{vy}^2/s_x^2
    \end{matrix} \right]
\end{equation}

Note the impact the X-location of the sensor on the vehicle has on the covariance. If the sensor's
X coordinate coincides with that of either ICR axis, then the related angular velocity cannot be observed.
Thus, for better accuracy it is advisable to place the sensor far from these axes along the X direction.

\section{EXPERIMENTS AND RESULTS}
\label{sec:experiments}

\begin{figure}
    \centering 
    \includegraphics[width=1\linewidth]{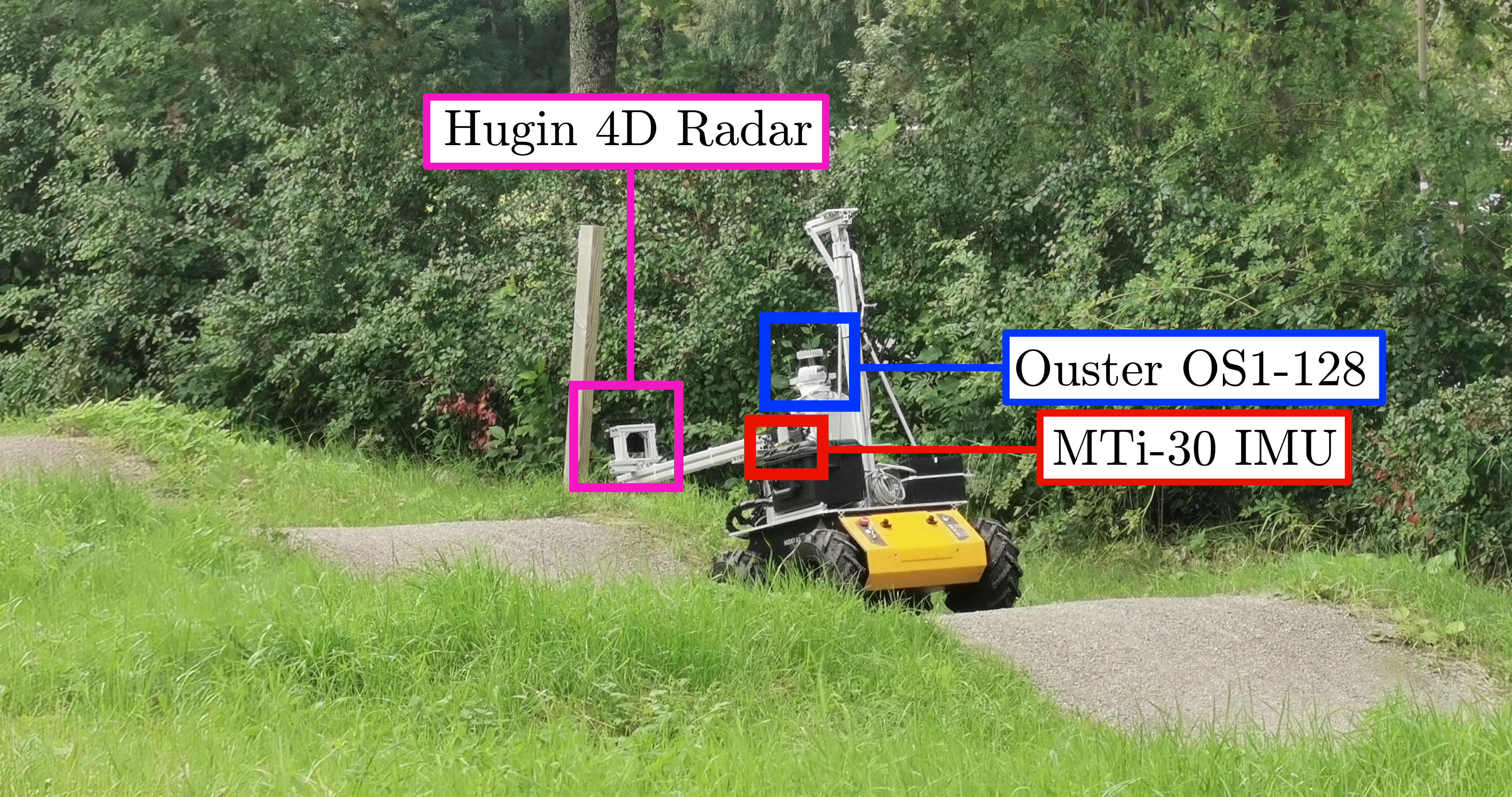}
    \caption{Sensor setup and test environment from sequence 04. 
    }
    \label{fig:husky}
\end{figure}

In this section, we describe the experiments performed in order to validate the proposed method, and subsequently evaluate it based on the obtained results.
The employed setup consists of a Hugin 4D imaging radar, an Ouster OS1-128 3D lidar, and an Xsens MTi-30 IMU all mounted on a Clearpath Husky robotic base. The proposed method makes use only of the radar, while the lidar and IMU are used here for comparison in the evaluation.
A dataset was collected while the Husky was driven around sloped terrain. 
Six different sequences were recorded: 04 and 05 have the most hills and thus the most vertical movement; 06 represents driving on a flat ground; 07 and 08 are similar multifaceted trajectories, with the second having a higher velocity throughout; and 11 is a longer route.
Both the environment and the sensor setup can be seen in Figure \ref{fig:husky}, and the dataset is publicly available at \href{https://zenodo.org/record/8346769}{https://zenodo.org/record/8346769}.
Note how the sensor is mounted on a lever arm on the vehicle. This is because increasing the distance along X from the sensor to both ICR axes results in a more accurate angular velocity, as established in Section \ref{sec:covariance}.
On a larger vehicle, like a car, the sensor can be mounted far from the ICR without needing a lever arm.

In contrast to the Ackermann vehicle used as an example in Section \ref{sec:kinematic}, the Husky is a skid-steer vehicle. 
As explained before, the proposed method works with every vehicle whose Z-ICR is in a consistent location.
We are aware of the implications of working with this type of vehicle \cite{skidsteer}, but the results show that the kinematic model is accurate enough to recover the angular velocity.

\subsection{SENSOR CALIBRATION}
\label{sec:calibration}

For the proposed method to properly work, the sensor orientation and position with respect to the vehicle must be known.
To that end, we propose a two-step calibration algorithm. 
First, the vehicle moves in a straight line, forward and backward. 
We then find the rotation that minimizes both Y and Z velocities calculated by the method, since only the X component should be non-zero.
The second step is similar, but in this case the rotation required is the one around the X axis, which is not determined by the previous step. Therefore, the vehicle moves freely around a flat surface, to then find the rotation that minimizes the estimated Z velocity.
By combining both rotations, the calculated sensor velocity can be transformed into the reference frame of the vehicle, and thus be used to obtain the angular velocities.

Just as important as the orientation of the sensor is its position, in particular, the distance along the X axis from the sensor to the location of the Z-ICR. This distance can be measured, but to avoid the possible error introduced by manually measuring it, we decided to calibrate $s_x$ employing the IMU data.
The procedure is simple: the vehicle moves in a flat circle at a constant speed, and then $s_x$ is chosen to minimize the difference between the calculated angular velocity and the one measured by the IMU.
Note that this is not the only possible way to calibrate the sensor position, but information about the movement is required either way.

\subsection{ACCURACY RESULTS}
\label{sec:results}

\begin{figure}[t]
  \centering
  \includegraphics[width=\linewidth]{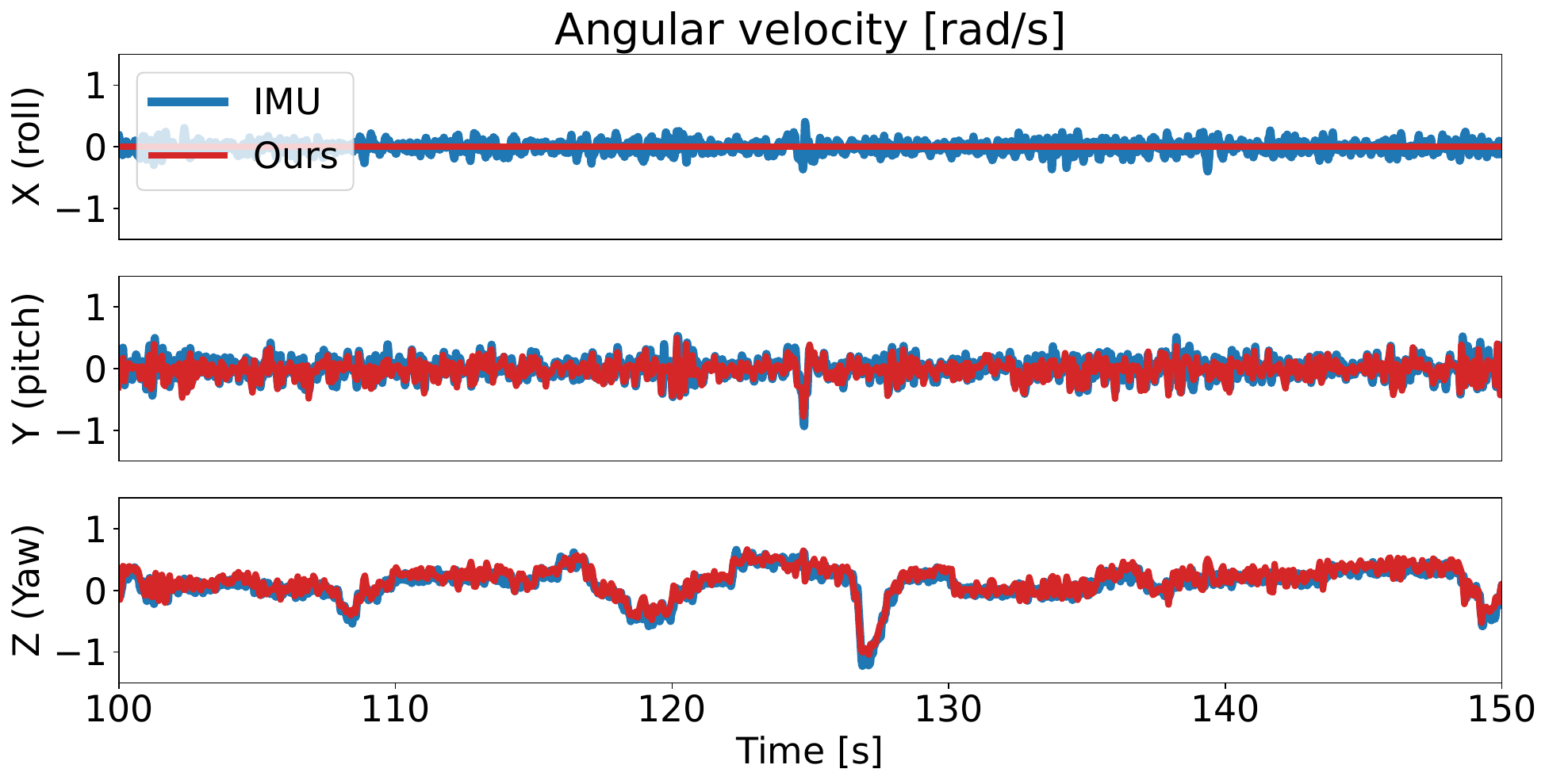}
  
  \caption{Angular velocity estimated from the Doppler velocities compared to IMU measurements, from sequence 11.}
  \label{fig:angvels}
\end{figure}

In this section we will evaluate the accuracy of the proposed method for providing single-scan and single-sensor 3D motion estimation. 
We compare the results to an IMU, since both provide a short-term accurate estimation of the motion, but show drift over time. As the ground truth for the comparison, we use the trajectory obtained by running an ICP-based mapper on the 3D lidar data. 

An IMU provides information on the angular velocity and linear acceleration of the sensor. Its trajectory can be estimated by integrating these values over time. 
However, these measurements suffer from having noise and a non-constant bias. Combined with the high frequency at which these sensors operate, said trajectory shows significant drift over time.
The position is impacted even more
since two integration steps are performed from the measured linear acceleration \cite{woodman2007introduction}. 
Therefore, the translation estimated from the IMU is not reliable enough to be included in the results.

Our method, in contrast to the IMU integration, obtains the translation of the sensor directly from its estimated velocity. 
The rotation, however, depends also on the calibration previously mentioned, as well as the fulfillment of the kinematic hypothesis. 
Our simple approach allows for fast execution, taking on average 4.50~$\pm$~1.94~ms per scan. 

Figure~\ref{fig:angvels} shows the angular velocities estimated by the proposed method compared to those measured by the IMU. Despite the mentioned limitations, our method provides an estimation similar to the IMU. 
Notice also how the roll velocity measured by the IMU is not as significant as the yaw and pitch. 
It is normal to undergo some roll rotation in the rough terrain of the experiments, but it is small enough to consider it inconsequential. On smoother surfaces, like roads, it will be even smaller.
This supports the restriction imposed by the kinematic model on the roll.

It makes sense to combine the advantages of both approaches into one, namely the translation from Doppler velocities with the rotation from the IMU, as done by Doer and Trommer \cite{DoerMFI2020}.
However, this combined method needs information from two different sensors to recover the motion. 

\begin{figure}
    \centering
    \includegraphics[trim={0 125px 0 0}, clip, width=1\linewidth]{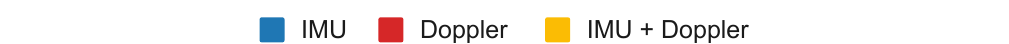}
    
    \includegraphics[width=1\linewidth]{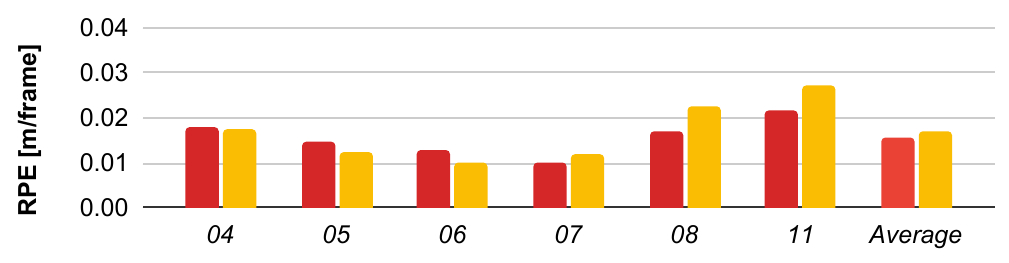}

    \includegraphics[ width=1\linewidth]{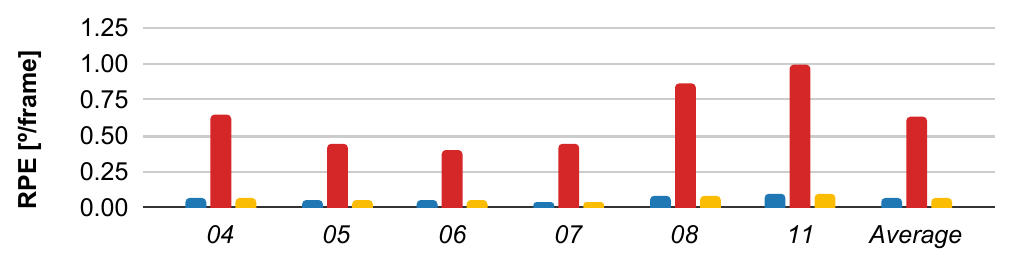}
    
    \caption{Translational and rotational RPE per frame.}
    \label{fig:rpef}
\end{figure}

To compare these three approaches (IMU, Doppler, IMU + Doppler), we include in Figure \ref{fig:rpef} the resulting Relative Pose Error (RPE) per frame of the test sequences. This evaluates the short-term accuracy of the different methods, without taking into account their drift over time.
Regarding the translation, IMU-only has been omitted because of its large error as discussed above, with the trajectory being hundreds of meters away from the ground truth.
Both Doppler-only and IMU + Doppler recover the translation in a similar manner, and thus it makes sense for them to provide comparable accuracy.
The difference is more apparent in the rotation. The accuracy of our method depends not only on the noise of the input data, unlike IMU. The fulfillment of the kinematic model and the calibration play an important role, hindering the rotation estimation.
Sequences 04, 08 and 11 stand out as the most challenging because of their uneven terrain, causing more vibrations and defying the no-roll assumption.

\subsection{INITIALIZATION OF SLAM}
\label{sec:slam}

The proposed method's short term accuracy and fast execution make it an interesting candidate to provide a motion prior to a more complex and accurate localization method, \eg SLAM.
ICP \cite{icp} is a widespread point cloud registration algorithm used among SLAM methods but is prone to falling into local minima. By providing a motion prior this limitation can be greatly reduced.
It is already fairly common in the literature to employ an IMU to initialize the motion. 

To evaluate the impact of using the proposed method as initialization for SLAM, we feed the radar data and a motion prior to an ICP-based mapper and analyze its accuracy based on how the motion is initialized. 
We test 3 different prior estimators: IMU only, Doppler only (our method), and IMU combined with radar. The trajectory of the mapper without any prior is also included as the base case in the comparison.
Again, the ground truth used for the comparison is the one obtained from running the mapper on the 3D lidar data.
The evaluation metric in this case is the RPE per second, which focuses more on long-term accuracy, a desirable trait for SLAM algorithms.

\begin{figure}
    \centering
    \includegraphics[trim={0 125px 0 0}, clip, width=1\linewidth]{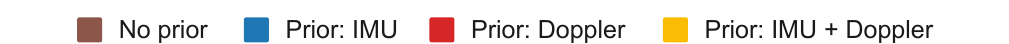}
    \includegraphics[width=1\linewidth]{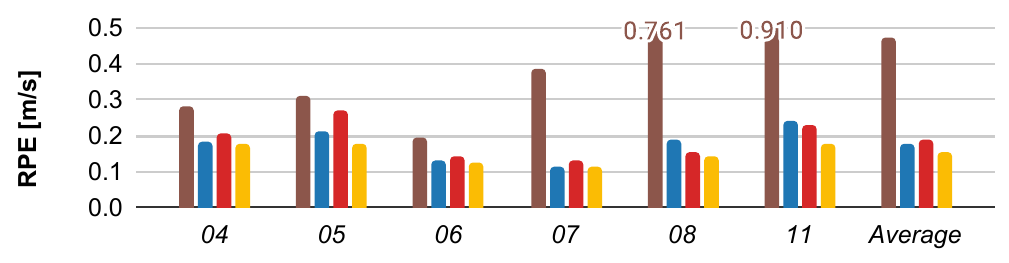}
    \includegraphics[width=1\linewidth]{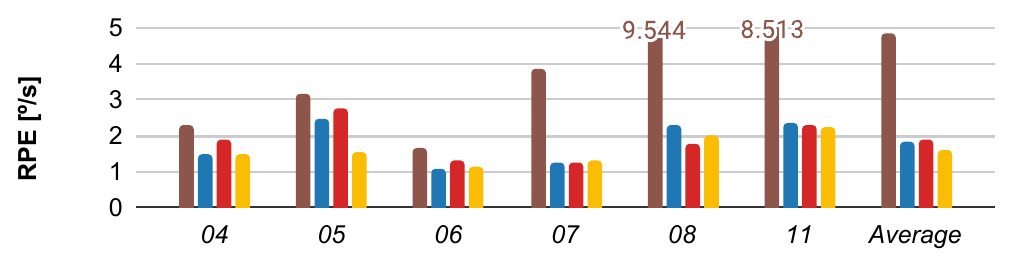}
    
    \caption{RPE per second of the mapper employing different prior data. Values out of the graph are labeled.}
    \label{fig:rpes}
\end{figure}

\begin{figure}
  \centering

  \includegraphics[width=1\linewidth]{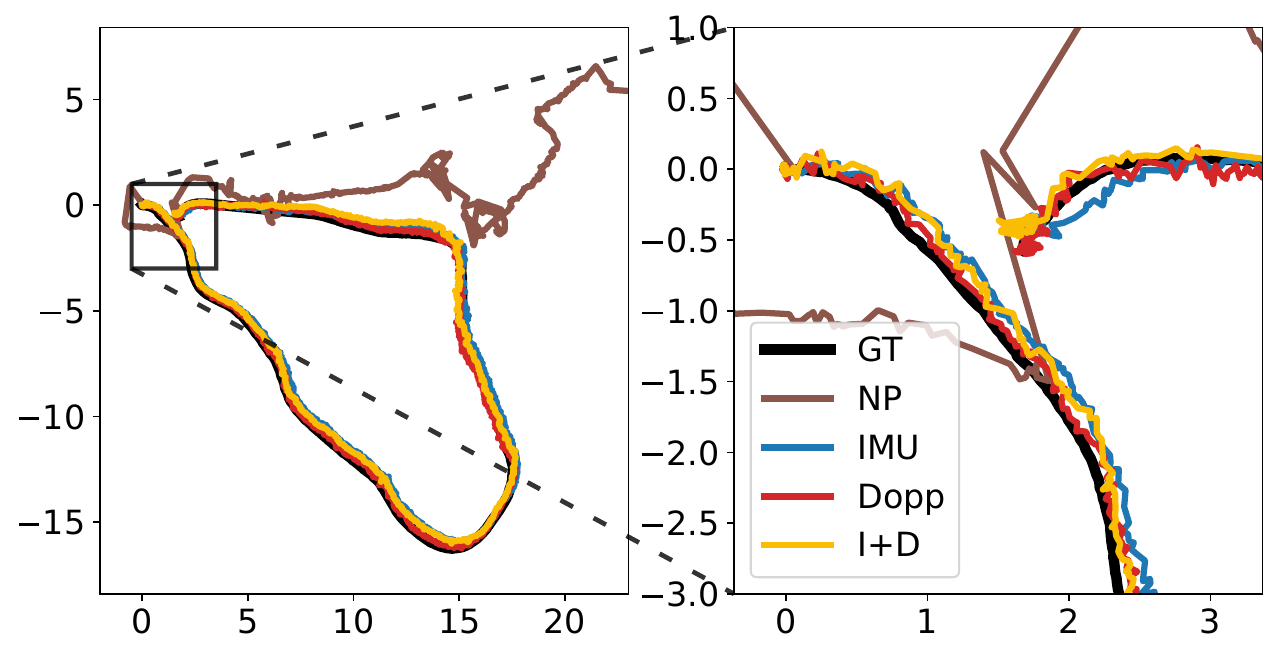}
  
  \caption{Trajectory in meters generated by the mapper with different prior data on sequence 07 (left), and a closer view of the start and finish points (right). 
  Prior used: none (NP), IMU, our method (Dopp), and IMU and Doppler (I+D).
  }
  \label{fig:traj}
\end{figure}

The resulting error values of the experiments can be seen in Figure \ref{fig:rpes}, and the trajectories in Figure \ref{fig:traj}.
Despite the variation along different sequences, all methods perform at a similar level of accuracy.
What is obvious is the need for a prior input to the mapper, without which it fails. 
Our method provides a reliable motion prior, resulting in a similar accuracy to IMU, without the need to mount and calibrate an extra sensor. Furthermore, the proposal can identify and remove dynamic points from the scene, which can help at later stages of the mapper pipeline.

\section{CONCLUSIONS}

In this paper, we have introduced a fast 3D odometry that estimates the motion of a vehicle from a single scan leveraging the Doppler velocity.
The method first obtains the sensor velocity based on the measured radial velocities of points from the scene, while rejecting dynamic objects.
Based on the kinematic model of the vehicle, the relation between its velocity and the sensor's is established, and thus the 3D motion of the vehicle can be recovered.

A series of experiments have been carried out to validate our method, evaluating the accuracy of the odometry both when used as-is for pose tracking and when used as input to an ICP-based mapping pipeline.
We compared its performance with an IMU, given their similar ability to efficiently provide short-term accurate odometry.
These qualities make the proposal especially suitable as initialization for radar-only SLAM algorithms, as it boosts the accuracy and identifies dynamic objects at a very low cost.
The results prove how our method provides a similar accuracy to an IMU without needing an external sensor mounted on the vehicle.
Both the code and the dataset are publicly available.

At the core of the method lies the kinematic model of the vehicle. 
Granted that it simplifies the movement enough to be observable with a single sensor, it also sets the limitations needed for the proposal to accurately estimate the motion.
Challenging scenarios include the Z-ICR being unstable, for example in slippery terrain, bumpy roads with abrupt and uneven vertical variation (potholes), and roads with considerable cant that generate roll rotation.
These issues cannot be solved solely through software.
On the other hand, the impact of having large amounts of outliers or noisy measurements on the accuracy can be reduced by correctly filtering the data, which should be the focus for future work.


\end{document}